\pdfoutput=1
\PassOptionsToPackage{table,xcdraw}{xcolor}
\documentclass[11pt]{article}

\usepackage[final]{acl}

\usepackage{times}
\usepackage{latexsym}

\usepackage[T1]{fontenc}

\usepackage[utf8]{inputenc}

\usepackage{microtype}

\usepackage{inconsolata}

\usepackage{graphicx}
\usepackage{subfigure}
\usepackage{bm}
\usepackage{amsthm}
\usepackage{amsmath,amsfonts}
\usepackage{bbding}
\usepackage{enumitem}
\usepackage{graphicx}
\usepackage{textcomp}
\usepackage{xcolor}
\newtheorem{theorem}{Theorem}

\usepackage[linesnumbered,ruled,vlined]{algorithm2e}
\SetKwInput{KwInput}{Input}               
\SetKwInput{KwOutput}{Output}
\usepackage{float}
\usepackage{booktabs}
\usepackage{graphicx}
\usepackage{multirow}
\usepackage[most]{tcolorbox}
\theoremstyle{remark}
\usepackage[export]{adjustbox}
\newtheorem{remark}[theorem]{Remark}
\newcommand{\name}{\textsf{BlendFilter}}

\usepackage{setspace}
\AtBeginDocument{%
  \addtolength\abovedisplayskip{-0.05\baselineskip}%
  \addtolength\belowdisplayskip{-0.05\baselineskip}%
  \addtolength\abovedisplayshortskip{-0.05\baselineskip}%
  \addtolength\belowdisplayshortskip{-0.05\baselineskip}%
}
\usepackage{titlesec}
\titlespacing*{\section}{0pt}{0.25\baselineskip}{0.25\baselineskip}
\titlespacing*{\subsection}{0pt}{0.25\baselineskip}{0.25\baselineskip}
\titlespacing*{\subsubsection}{0pt}{0.25\baselineskip}{0.25\baselineskip}
\title{\includegraphics[valign=b, width=0.6cm]{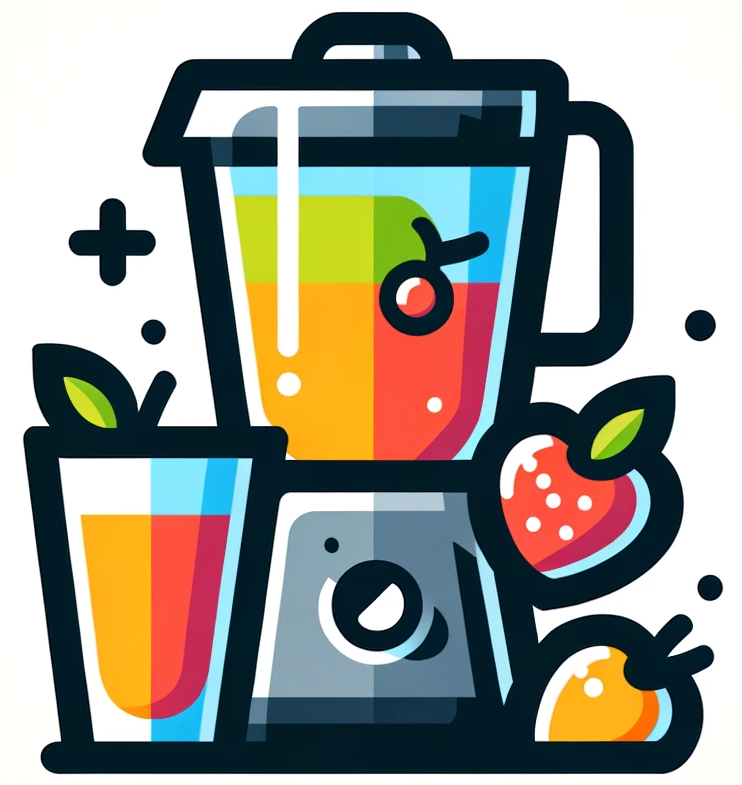}{\name}: Advancing Retrieval-Augmented Large Language Models via Query Generation Blending and Knowledge Filtering}

\author{Haoyu Wang$^\dagger{}$, Ruirui Li$^\star{}$, Haoming Jiang$^\star{}$, Jinjin Tian$^\star{}$, Zhengyang Wang$^\star{}$, Chen Luo$^\star{}$,\\ \textbf{Xianfeng Tang}$^\star{}$, \textbf{Monica Xiao Cheng}$^\star{}$, \textbf{Tuo Zhao}$^*$, \textbf{Jing Gao}$^{\S}$ \\
  $^\dagger{}$SUNY Albany,
  $^{\S}$Purdue University,
  $^*$Georgia Institute of Technology, 
  $^\star{}$Amazon\\
  \texttt{$^\dagger{}$hwang28@albany.edu},
  \texttt{$^{\S}$jinggao@purdue.edu}, \texttt{$^*$tourzhao@gatech.edu} , \\\texttt{$^\star{}$\{ruirul,jhaoming,jinjint,zhengywa,cheluo,xianft,chengxca\}@amazon.com}}

\begin{document}
\maketitle
\begin{abstract}
Retrieval-augmented Large Language Models (LLMs) offer substantial benefits in enhancing performance across knowledge-intensive scenarios. However, these methods often face challenges with complex inputs and encounter difficulties due to noisy knowledge retrieval, notably hindering model effectiveness. To address this issue, we introduce {\name}, a novel approach that elevates retrieval-augmented LLMs by integrating query generation blending with knowledge filtering. {\name} proposes the blending process through its query generation method, which integrates both external and internal knowledge augmentation with the original query, ensuring comprehensive information gathering. Additionally, our distinctive knowledge filtering module capitalizes on the intrinsic capabilities of the LLM, effectively eliminating extraneous data. We conduct extensive experiments on three open-domain question answering benchmarks, and the findings clearly indicate that our innovative {\name} surpasses state-of-the-art baselines significantly.
\end{abstract}

\section{Introduction}
Generative Large Language Models (LLMs) have shown remarkable proficiency in various applications, such as summarization~\cite{zhang2023benchmarking,wang2023zero}, dialogue systems~\cite{hudevcek2023large,touvron2023llama}, and question answering~\cite{lazaridou2022internet,lu2022learn}. Nonetheless, the finite scope of their pre-training corpora imposes inherent limitations, preventing LLMs from capturing and maintaining comprehensive worldly knowledge, especially given its dynamic nature. This limitation has spurred interest in retrieval-augmented generation strategies that integrate external knowledge sources, like Wikipedia, to refine the quality of LLM-generated content.

Typically, retrieval-augmented generation methods~\cite{brown2020language,izacard2022few,zakka2023almanac} feed a task input, such as a user query or a question in open-domain question answering, into a retriever to obtain related knowledge documents. Subsequently, the LLM generates content based on the initial input and the information retrieved. Nevertheless, this direct retrieval strategy faces challenges with intricate task inputs~\cite{shao-etal-2023-enhancing}. While straightforward queries enable effective identification of relevant information, multifaceted and complex questions may not cover some essential  keywords, complicating the retrieval of pertinent documents.

To enhance the retrieval for complex task inputs, recent studies have proposed methods to enrich the original input. These approaches encompass question decomposition~\cite{yao2022react, press2022measuring}, query rewriting~\cite{ma2023query}, and query augmentation~\cite{yu2023improving, shao-etal-2023-enhancing}. They utilize knowledge memorized by LLMs or sourced from external databases to supplement the input with additional information, thereby explicitly incorporating additional keywords and substantially facilitating the retrieval process. Among these, query augmentation is particularly noteworthy and achieves state-of-the-art performance because it processes all retrieved knowledge collectively while generating answers and it does not require the training of an additional language model for query rewriting. 

However, current query augmentation methods still suffer from some limitations. These techniques have typically relied on a single source of augmentation, either LLM internal knowledge or an external knowledge base. On one hand, for certain complex inputs, this single source of augmentation may not be able to cover all the keywords and thus lead to insufficient augmentation. 
Furthermore, existing work excludes original input but only rely on the augmented query, which could further exacerbate information loss. 

Another major problem of existing methods is that the incorporated content fetched by the retriever could contain irrelevant or misleading information. Usually top-$K$ returned documents by the retriever will be used as augmentation, but there is no guarantee that all the top-$K$ documents are relevant and helpful for the task. 
Correspondingly, incorporating such noise information into the augmented query can potentially lead to inaccuracies in the LLM's output~\cite{wang2023self}. To mitigate the noise in retrieved knowledge documents, previous studies~\cite{yu2023improving, wang2023self, asai2023self} have suggested various strategies. Unfortunately, these existing noise reduction methods in knowledge document retrieval are dependent on the LLM's confidence levels, which can be imprecise~\cite{xiong2023can}. Additionally, these methods often require an extra language model to determine the need for retrieval, which incurs significant computational costs.

\begin{figure*} 
\centering        
\includegraphics[width=1.9\columnwidth]{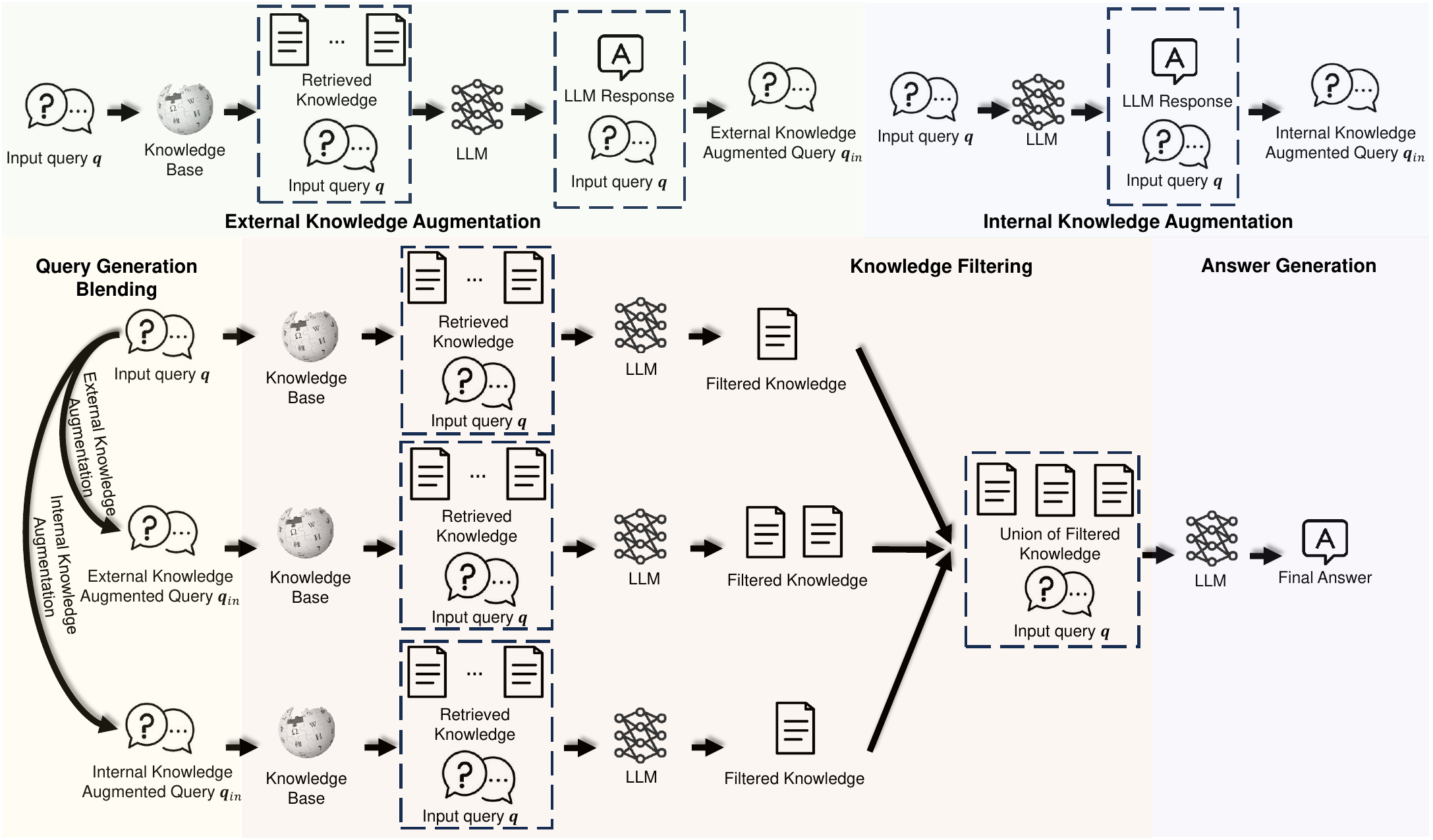}
\caption{The framework of {\name}.} 
\label{fig:framework}
\vspace{-0.1in}
\end{figure*}

To tackle the aforementioned \textit{complex question} and \textit{noisy retrieved knowledge} challenges, we propose \includegraphics[valign=b, width=0.5cm]{icon.png} {\name}, a novel framework that advances retrieval-augmented large language models by integrating query generation blending and knowledge filtering, as illustrated in Fig.~\ref{fig:framework}. Our framework, {\name}, is structured around three core components: 1) \textbf{Query Generation Blending} module, 2) \textbf{Knowledge Filtering} module, and a  3) Answer Generation module. The \textbf{Query Generation Blending} module is dedicated to enhancing input queries through diverse augmentation strategies, essentially forming a composite of queries, to handle the complex question challenge. This module incorporates both external and internal knowledge sources for augmentation. These augmented queries, including the original, external knowledge-augmented, and internal knowledge-augmented, are then employed by the retriever to collect pertinent information.
In order to tackle the noise retrieved knowledge challenge, our proposed \textbf{Knowledge Filtering} module, aims to eliminate irrelevant retrieved knowledge and could operate autonomously without needing an extra language model, leveraging the innate filtering prowess of the LLM. In the final phase, the LLM integrates the filtered knowledge with the original query to generate the final answer.

The contributions are summarized as follows: 1)~We introduce a novel query generation blending approach that integrates various augmentation sources. In contrast to existing work that relies on one source only, the proposed method enriches queries by using a variety of knowledge sources, which lead to a more comprehensive coverage of pertinent knowledge. 2)~We present a novel and effective knowledge filtering module designed to eliminate irrelevant knowledge. We are the first to propose the utilization of the LLM itself as a filter in retrieval-augmented generation tasks. 3)~We conduct extensive experiments across three open-domain question answering benchmarks. The results demonstrate that our proposed model, {\name}, significantly surpasses the baseline models across three distinct backbones.

\section{Related Work}

Retrieval-augmented generation enhances Large Language Models (LLMs) by leveraging external knowledge to improve generation quality. Initial approaches, as discussed in \cite{izacard2021leveraging,shao2021answering,izacard2022atlas,shi2023replug}, portrayed LLMs as passive recipients of retrieved knowledge, lacking interactive dynamics with retrievers. However, due to the inherent challenges in accurately capturing relevance between inputs and documents, these direct methods often yield only marginal improvements. Addressing this, recent advancements \cite{nakano2021webgpt,trivedi2022interleaving,jiang2023active,li2023agent4ranking,li2023llatrieval,wang2023self,asai2023self,yu2023improving,ma2023query,press2022measuring,yao2022react} have empowered LLMs to engage actively with retrievers, thereby enhancing relevance modeling. The integration of LLMs into the retrieval process broadly falls into three categories: 1) question decomposition, 2) query rewriting, and 3) query augmentation. For question decomposition, as exemplified by \citet{yao2022react} and \citet{press2022measuring}, LLMs break down a complex question into simpler components, leveraging both previous interactions and retrieved knowledge. This decomposition facilitates more straightforward reasoning by LLMs. However, the success of this approach heavily depends on the LLM's capabilities. Insufficiently powerful LLMs might generate misleading sub-questions. Moreover, this method requires maintaining a historical context, potentially leading to lengthy dialogues and increased computational costs. In the realm of query rewriting, models are trained, often utilizing reinforcement learning, to reformulate the original question into a version more conducive to retrieval \cite{ma2023query,li2023agent4ranking}. These revised questions typically yield improved generation outcomes. Nevertheless, training an additional model for rewriting is a resource-intensive process. The third approach, query augmentation, involves enriching queries with knowledge from either LLM internal databases or external sources \cite{shao-etal-2023-enhancing,yu2023improving}. A limitation of this method is its reliance on a single source of augmentation and often overlooking the original query, thus constraining overall model performance.

The aforementioned studies directly utilize retrieved knowledge, yet recent research~\cite{wang2023self,li2023llatrieval} highlights that such knowledge can sometimes be irrelevant or even detrimental to LLMs when answering queries. To solve this challenge, \cite{wang2023self} suggests an initial assessment to determine if LLMs need to retrieve knowledge, utilizing a classifier that could be based on BERT-like models or the LLM itself. However, this approach requires additional training data, which poses challenges in zero-shot or few-shot learning scenarios, and the LLM's self-evaluation may not always yield reliable results. \cite{asai2023self} introduces a self-reflective method to ascertain the necessity of retrieval and to assess the relevance between the retrieved knowledge and the input. A critical limitation of this method, as noted by \cite{asai2023self}, is its dependence on training an auxiliary language model to produce text with reflection tokens, incurring extra costs. Additionally, \cite{yu2023improving} employs a strategy of comparing the average negative likelihood of answers with and without external knowledge to guide decision-making. Nevertheless, this measure may not be a precise indicator of model confidence and is not universally applicable across models, with certain models like GPT-3.5-turbo and GPT-3.5-turbo-Instruct currently unable to access this feature.
We summarize the \textbf{differences} between the proposed {\name} and other baselines in Table~\ref{tab:diffs} in the appendix.

\section{Methodology} 
Given a pre-trained Large Language Model (LLM) $\mathcal{M}(\cdot)$, a knowledge base $\mathcal{K}=\{\mathcal{K}_{i}\}_{i=1}^{n}$ (where $n$ represents the number of documents), a retriever $\mathcal{R}(\cdot)$, and a query $\bm{q}$, our objective is to utilize the knowledge base to facilitate accurate responses from the LLM without fine-tuning.
\subsection{Overview}
To enhance the retrieval quality for retrieval-augmented LLMs, we introduce a framework named {\name}, which incorporates query generation blending and knowledge filtering, as depicted in Fig.~\ref{fig:framework}. We begin by presenting query blending, a technique that enhances the original query by incorporating both external knowledge and the LLM's internally memorized knowledge (Section~\ref{sec:blending}). Additionally, we propose a knowledge filtering module to effectively remove irrelevant knowledge (Section~\ref{sec:filtering}). Finally, we demonstrate how the LLM generates answers based on the filtered knowledge (Section~\ref{sec:generation}).
\subsection{Query Generation Blending}
\label{sec:blending}
Numerous studies~\cite{izacard2021leveraging, shao2021answering, izacard2022atlas, shi2023replug} have validated the effectiveness of utilizing a retriever to enrich questions with relevant knowledge, thereby boosting the performance of LLMs. This process can be represented as follows: $\mathcal{K}_{r}=\mathcal{R}(\bm{q},\mathcal{K};K)$, $\bm{a}=\mathcal{M}(\bm{a}|\texttt{Prompt}(\bm{q},\mathcal{K}_{r}))$,
where $\bm{a}$ represents the generated answer, $\mathcal{K}_{r}$ denotes the retrieved knowledge, and $K$ serves as the hyper-parameter for the retriever, controlling the quantity of retrieved knowledge items. Nonetheless, in cases where the query is complex, directly inputting it into the retriever often fails to retrieve the correct knowledge documents. As a solution, we advocate for the incorporation of both external and internal knowledge augmentation techniques to refine the query.\\
\noindent\textbf{External Knowledge Augmentation.} For complex questions, such as those in multi-hop question answering~\cite{yang2018hotpotqa}, which often entail implicit sub-problems and span multiple knowledge domains, we utilize an external knowledge base to refine the original query and facilitate document retrieval. Specifically, we initially retrieve relevant knowledge documents using the original query, as follows: $\mathcal{K}_{ex}=\mathcal{R}(\bm{q},\mathcal{K};K).$

Subsequently, we engage the LLM to derive the answer using the acquired knowledge documents via the Chain-of-Thought (CoT) approach~\cite{wei2022chain}. This step is  depicted as: $\bm{a}_{ex}=\mathcal{M}(\bm{a}|\texttt{Prompt}_\texttt{CoT}(\bm{q},\mathcal{K}_{ex}))$,
where $\bm{a}_{ex}$ represents the reasoning and answer generated by the LLM based on the retrieved knowledge $\mathcal{K}_{ex}$. 
The generated context $\bm{a}_{ex}$ contains related keywords and valuable information through CoT reasoning based on retrieved knowledge from the external knowledge base, thereby assisting the retriever in pinpointing relevant knowledge. Subsequently, we integrate the generated context $\bm{a}_{ex}$ with the initial query $\bm{q}$ to formulate the enhanced query, as shown below: $\bm{q}_{ex}=\bm{a}_{ex}\|\bm{q},$
where $\|$ represents the concatenation operation.
\begin{remark}
    This process of external knowledge augmentation essentially acts as a two-hop reasoning mechanism to refine the query. In fact, it can be extended to higher-order augmentation, but typically, leveraging two-hop information proves to be sufficiently effective in enhancing retrieval accuracy due to the LLM's strong capabilities. Consequently, we refrain from employing higher-order augmentation in order to strike a balance between efficiency and accuracy.
\end{remark}
\noindent\textbf{Internal Knowledge Augmentation.} LLMs have memorized a lot of factual knowledge. Some related knowledge is not retrieved in external knowledge augmentation while LLMs may memorize them internally. Consequently, we can prompt the LLM to produce a detailed response to the query, drawing upon its internal knowledge. This internally-sourced response acts as a supplement to the external knowledge. Specifically, the generated text based on LLM internal knowledge can be formulated as $\bm{a}_{in}=\mathcal{M}(\bm{a}|\texttt{Prompt}(\bm{q}))$,
and the augmented query is $\bm{q}_{in}=\bm{a}_{in}\|\bm{q}$.

\subsection{Knowledge Filtering}
\label{sec:filtering}
By integrating both external and internal knowledge-augmented queries in conjunction with the original query, we are able to retrieve the corresponding knowledge documents separately, as follows: $\mathcal{K}_{q}=\mathcal{R}(\bm{q},\mathcal{K};K)$, $\mathcal{K}_{q_{ex}}=\mathcal{R}(\bm{q}_{ex},\mathcal{K};K)$, $\mathcal{K}_{q_{in}}=\mathcal{R}(\bm{q}_{in},\mathcal{K};K)$,
where $\mathcal{K}_{q}$ represents knowledge documents retrieved by the original query, $\mathcal{K}_{q_{ex}}$ corresponds to the external knowledge-augmented query, and $\mathcal{K}_{q_{in}}$ pertains to the internal knowledge-augmented query. A direct approach to leveraging this retrieved knowledge involves taking their union: $\mathcal{K}_{r}^{direct}=\mathcal{K}_{q}\bigcup \mathcal{K}_{q_{ex}} \bigcup \mathcal{K}_{q_{in}}$.

This method ensures that the synthesized knowledge, $\mathcal{K}_{r}^{direct}$, encompasses a broader spectrum of relevant documents, thereby enhancing the quality of the retrieved knowledge. Nonetheless, retrieving some unrelated documents is inevitable due to the inherent imperfections of the retrieval process and the selection of the top-$K$ documents, which may include irrelevant information when $K$ exceeds the number of ground truth knowledge documents. This unrelated information can potentially lead to confusion and misguidance for the LLM, resulting in incorrect outputs. Rather than training a separate knowledge filter to identify and eliminate unrelated information, we have observed that the LLM itself serves as an effective knowledge filter. We provide both the original query and the retrieved knowledge to the Large Language Model (LLM) and instruct the LLM to perform knowledge filtering. This can be formulated as follows: $\mathcal{K}_{q}^{f}=\mathcal{M}(\mathcal{K}|\texttt{Prompt}(\bm{q},\mathcal{K}_{q}))$, $\mathcal{K}_{q_{ex}}^{f}=\mathcal{M}(\mathcal{K}|\texttt{Prompt}(\bm{q},\mathcal{K}_{q_{ex}}))$, $\mathcal{K}_{q_{in}}^{f}=\mathcal{M}(\mathcal{K}|\texttt{Prompt}(\bm{q},\mathcal{K}_{q_{in}}))$.

The final knowledge utilized for generation is obtained by taking the union of the filtered knowledge sets, i.e. $\mathcal{K}_{r}=\mathcal{K}_{q}^{f}\bigcup \mathcal{K}_{q_{ex}}^{f} \bigcup \mathcal{K}_{q_{in}}^{f}$,
where $\bigcup$ represents taking union operation.
\begin{remark}
    Our method involves filtering knowledge and subsequently combining the filtered information. An alternative option is to reverse the sequence of these two steps. However, we have observed that commencing with the union of knowledge may result in a larger knowledge set, consequently intensifying the challenge of subsequent knowledge filtering. Consequently, we opt to filter knowledge independently for $\mathcal{K}_{q}$, $\mathcal{K}_{q_{ex}}$, and $\mathcal{K}_{q_{in}}$.
\end{remark}
\subsection{Answer Generation}
\label{sec:generation}
In this step, the LLM generates an answer based on both the filtered knowledge and the original query. We employ CoT to enhance the model's reasoning performance, a representation of which is as follows: $\bm{a}=\mathcal{M}(\bm{a}|\texttt{Prompt}_{\texttt{CoT}}(\bm{q},\mathcal{K}_{r}))$.
The whole algorithm is summarized in Algorithm~\ref{alg:rag} in the appendix.

\section{Experiment}
In this section, we evaluate the proposed {\name} and answer the following research questions: \textbf{RQ1})~How does {\name} perform compared to state-of-the-art retrieval-augmented baselines? \textbf{RQ2})~Can the proposed {\name} generalize well with respect
to different backbones and retrievers? \textbf{RQ3})~Is the LLM effective to filter unrelated knowledge documents? \textbf{RQ4})~What are the roles of the original query, external knowledge-augmented query, and internal knowledge-augmented query in model performance improvements respectively? \textbf{RQ5})~How does the performance change with varying numbers of knowledge documents? \textbf{RQ6})~Will the proposed {\name} be improved by sampling multiple times with different temperatures?

\subsection{Datasets and Experiment Settings}
\subsubsection{Datasets}We conduct experiments on three public benchmarks, including HotPotQA~\cite{yang2018hotpotqa}, 2WikiMultiHopQA~\cite{ho2020constructing}, and StrategyQA~\cite{geva2021did}. Examples are illustrated in Fig.~\ref{fig:data} in the appendix.

\subsubsection{Evaluation Metrics} Following \citet{shao-etal-2023-enhancing}, we evaluate the first 500 questions from the training dataset for StrategyQA and 500 questions from the development dataset for HotPotQA and 2WikiMultiHopQA. For multi-hop question answering datasets, we employ exact match~(EM) and F1 as evaluation metrics, and for the commonsense reasoning dataset, we use accuracy, following \citet{yao2022react} and \citet{shao-etal-2023-enhancing}. To evaluate the retrieval performance, we leverage widely used Recall and Precision as evaluation metrics. Additionally, to assess the effectiveness of the proposed knowledge filtering in eliminating irrelevant information, we introduce a new metric called S-Precision. This metric measures the proportion of questions for which the retrieved documents precisely match the golden relevant documents.
\subsubsection{Baselines}We adopt following state-of-the-art baselines to evaluate our proposed {\name}: 1)~Direct Prompting~\cite{brown2020language}, 2)~CoT Prompting~\cite{wei2022chain}, 3)~ReAct~\cite{yao2022react}, 4)~SelfAsk~\cite{press2022measuring}, and 5)~ITER-RETGEN~\cite{shao-etal-2023-enhancing}. We show the detail information about these baselines in the appendix.
\subsubsection{Implementation Details.}
We evaluate models with three different LLMs: GPT3.5-turbo-Instruct$\footnote{\url{https://platform.openai.com/docs/models/gpt-3-5}}$, Vicuna 1.5-13b~\cite{zheng2023judging}, and Qwen-7b~\cite{qwen}. We utilize the state-of-the-art efficient retrieval method ColBERT v2~\cite{santhanam2022colbertv2} as the retriever implemented by \citet{khattab2022demonstrate,khattab2023dspy}. The knowledge base we employ is the collection of Wikipedia abstracts dumped in 2017~\cite{khattab2023dspy}. We show the detailed information about implementation details in the appendix.


\subsection{Performance Comparison}

In this section, we evaluate the performance of both the baseline models and our proposed {\name} model using various backbones. The results are displayed in Table~\ref{tab:gpt-turbo}, Table~\ref{tab:vicuna}, and Table~\ref{tab:qwen}, addressing \textbf{RQ1} and \textbf{RQ2}.

\begin{table*}[htb!]
\centering
\caption{Performance of {\name} with GPT3.5-turbo-Instruct as the backbone. IMP represents the percentage of improvements compared to baselines with respect to Exact Match on HotPotQA and 2WikiMultihopQA and Accuracy on StrategyQA.}
\vspace{-0.1in}
\label{tab:gpt-turbo}
\resizebox{.8\textwidth}{!}{%
\begin{tabular}{@{}lcccccccc@{}}
\toprule
\multicolumn{1}{c}{}                         & \multicolumn{3}{c}{\cellcolor[HTML]{FCF0EF}HotPotQA}                                                 & \multicolumn{3}{c}{\cellcolor[HTML]{F5F5FF}2WikiMultihopQA}                                          & \multicolumn{2}{c}{\cellcolor[HTML]{F8F2E9}StrategyQA}            \\ \cmidrule(l){2-9} 
\multicolumn{1}{c}{\multirow{-2}{*}{Method}} & \cellcolor[HTML]{FCF0EF}Exact Match & \cellcolor[HTML]{FCF0EF}F1    & \cellcolor[HTML]{FCF0EF}IMP    & \cellcolor[HTML]{F5F5FF}Exact Match & \cellcolor[HTML]{F5F5FF}F1    & \cellcolor[HTML]{F5F5FF}IMP    & \cellcolor[HTML]{F8F2E9}Accuracy & \cellcolor[HTML]{F8F2E9}IMP    \\ \midrule
\multicolumn{9}{c}{Without Retrieval}                                                                                                                                                                                                                                                                                          \\ \midrule
Direct                                       & \cellcolor[HTML]{FCF0EF}0.304       & \cellcolor[HTML]{FCF0EF}0.410 & \cellcolor[HTML]{FCF0EF}67.1\% & \cellcolor[HTML]{F5F5FF}0.282       & \cellcolor[HTML]{F5F5FF}0.318 & \cellcolor[HTML]{F5F5FF}43.3\% & \cellcolor[HTML]{F8F2E9}0.648    & \cellcolor[HTML]{F8F2E9}14.8\% \\
CoT                                          & \cellcolor[HTML]{FCF0EF}0.302       & \cellcolor[HTML]{FCF0EF}0.432 & \cellcolor[HTML]{FCF0EF}68.2\% & \cellcolor[HTML]{F5F5FF}0.300       & \cellcolor[HTML]{F5F5FF}0.403 & \cellcolor[HTML]{F5F5FF}34.7\% & \cellcolor[HTML]{F8F2E9}0.700    & \cellcolor[HTML]{F8F2E9}6.3\%  \\ \midrule
\multicolumn{9}{c}{With Retrieval}                                                                                                                                                                                                                                                                                             \\ \midrule
Direct                                       & \cellcolor[HTML]{FCF0EF}0.412       & \cellcolor[HTML]{FCF0EF}0.537 & \cellcolor[HTML]{FCF0EF}23.3\% & \cellcolor[HTML]{F5F5FF}0.318       & \cellcolor[HTML]{F5F5FF}0.371 & \cellcolor[HTML]{F5F5FF}27.0\% & \cellcolor[HTML]{F8F2E9}0.634    & \cellcolor[HTML]{F8F2E9}17.4\% \\
CoT                                          & \cellcolor[HTML]{FCF0EF}0.434       & \cellcolor[HTML]{FCF0EF}0.558 & \cellcolor[HTML]{FCF0EF}17.1\% & \cellcolor[HTML]{F5F5FF}0.318       & \cellcolor[HTML]{F5F5FF}0.396 & \cellcolor[HTML]{F5F5FF}27.0\% & \cellcolor[HTML]{F8F2E9}0.616    & \cellcolor[HTML]{F8F2E9}20.8\% \\
ReAct                                        & \cellcolor[HTML]{FCF0EF}0.360       & \cellcolor[HTML]{FCF0EF}0.475 & \cellcolor[HTML]{FCF0EF}41.1\% & \cellcolor[HTML]{F5F5FF}0.374       & \cellcolor[HTML]{F5F5FF}0.450 & \cellcolor[HTML]{F5F5FF}8.0\%  & \cellcolor[HTML]{F8F2E9}0.658    & \cellcolor[HTML]{F8F2E9}13.1\% \\
SelfAsk                                      & \cellcolor[HTML]{FCF0EF}0.364       & \cellcolor[HTML]{FCF0EF}0.481 & \cellcolor[HTML]{FCF0EF}39.6\% & \cellcolor[HTML]{F5F5FF}0.334       & \cellcolor[HTML]{F5F5FF}0.416 & \cellcolor[HTML]{F5F5FF}21.0\% & \cellcolor[HTML]{F8F2E9}0.638    & \cellcolor[HTML]{F8F2E9}16.6\%   \\
ITER-RETGEN                                  & \cellcolor[HTML]{FCF0EF}0.450       & \cellcolor[HTML]{FCF0EF}0.572 & \cellcolor[HTML]{FCF0EF}12.9\% & \cellcolor[HTML]{F5F5FF}0.328       & \cellcolor[HTML]{F5F5FF}0.436 & \cellcolor[HTML]{F5F5FF}23.2\% & \cellcolor[HTML]{F8F2E9}0.692    & \cellcolor[HTML]{F8F2E9}7.5\%  \\
BlendFilter                                  & \cellcolor[HTML]{FCF0EF}0.508       & \cellcolor[HTML]{FCF0EF}0.624 & \cellcolor[HTML]{FCF0EF}-      & \cellcolor[HTML]{F5F5FF}0.404       & \cellcolor[HTML]{F5F5FF}0.470 & \cellcolor[HTML]{F5F5FF}-      & \cellcolor[HTML]{F8F2E9}0.744    & \cellcolor[HTML]{F8F2E9}-      \\ \bottomrule
\end{tabular}%
}
\end{table*}

\begin{table*}[htb!]
\centering
\caption{Performance of {\name} with Vicuna 1.5-13b as the backbone.}
\vspace{-0.1in}
\label{tab:vicuna}
\resizebox{.8\textwidth}{!}{%
\begin{tabular}{@{}lcccccccc@{}}
\toprule
\multicolumn{1}{c}{}                         & \multicolumn{3}{c}{\cellcolor[HTML]{FCF0EF}HotPotQA}                                                 & \multicolumn{3}{c}{\cellcolor[HTML]{F5F5FF}2WikiMultihopQA}                                           & \multicolumn{2}{c}{\cellcolor[HTML]{F8F2E9}StrategyQA}            \\ \cmidrule(l){2-9} 
\multicolumn{1}{c}{\multirow{-2}{*}{Method}} & \cellcolor[HTML]{FCF0EF}Exact Match & \cellcolor[HTML]{FCF0EF}F1    & \cellcolor[HTML]{FCF0EF}IMP    & \cellcolor[HTML]{F5F5FF}Exact Match & \cellcolor[HTML]{F5F5FF}F1     & \cellcolor[HTML]{F5F5FF}IMP    & \cellcolor[HTML]{F8F2E9}Accuracy & \cellcolor[HTML]{F8F2E9}IMP    \\ \midrule
\multicolumn{9}{c}{Without Retrieval}                                                                                                                                                                                                                                                                                           \\ \midrule
Direct                                       & \cellcolor[HTML]{FCF0EF}0.202       & \cellcolor[HTML]{FCF0EF}0.267 & \cellcolor[HTML]{FCF0EF}96.0\% & \cellcolor[HTML]{F5F5FF}0.246       & \cellcolor[HTML]{F5F5FF}0.288  & \cellcolor[HTML]{F5F5FF}16.3\% & \cellcolor[HTML]{F8F2E9}0.604    & \cellcolor[HTML]{F8F2E9}11.3\% \\
CoT                                          & \cellcolor[HTML]{FCF0EF}0.228       & \cellcolor[HTML]{FCF0EF}0.344 & \cellcolor[HTML]{FCF0EF}73.7\% & \cellcolor[HTML]{F5F5FF}0.190       & \cellcolor[HTML]{F5F5FF}0.279  & \cellcolor[HTML]{F5F5FF}50.5\% & \cellcolor[HTML]{F8F2E9}0.660    & \cellcolor[HTML]{F8F2E9}1.8\%  \\ \midrule
\multicolumn{9}{c}{With Retrieval}                                                                                                                                                                                                                                                                                              \\ \midrule
Direct                                       & \cellcolor[HTML]{FCF0EF}0.336       & \cellcolor[HTML]{FCF0EF}0.443 & \cellcolor[HTML]{FCF0EF}17.9\% & \cellcolor[HTML]{F5F5FF}0.210       & \cellcolor[HTML]{F5F5FF}0.284  & \cellcolor[HTML]{F5F5FF}36.2\% & \cellcolor[HTML]{F8F2E9}0.624    & \cellcolor[HTML]{F8F2E9}7.7\%  \\
CoT                                          & \cellcolor[HTML]{FCF0EF}0.362       & \cellcolor[HTML]{FCF0EF}0.488 & \cellcolor[HTML]{FCF0EF}9.4\%  & \cellcolor[HTML]{F5F5FF}0.206       & \cellcolor[HTML]{F5F5FF}0.302  & \cellcolor[HTML]{F5F5FF}38.8\% & \cellcolor[HTML]{F8F2E9}0.646    & \cellcolor[HTML]{F8F2E9}4.0\%  \\
ReAct                                        & \cellcolor[HTML]{FCF0EF}0.332       & \cellcolor[HTML]{FCF0EF}0.463 & \cellcolor[HTML]{FCF0EF}19.3\% & \cellcolor[HTML]{F5F5FF}0.216       & \cellcolor[HTML]{F5F5FF}0.323  & \cellcolor[HTML]{F5F5FF}32.4\% & \cellcolor[HTML]{F8F2E9}0.588    & \cellcolor[HTML]{F8F2E9}14.3\% \\
SelfAsk                                      & \cellcolor[HTML]{FCF0EF}0.361       & \cellcolor[HTML]{FCF0EF}0.469 & \cellcolor[HTML]{FCF0EF}9.7\%  & \cellcolor[HTML]{F5F5FF}0.250       & \cellcolor[HTML]{F5F5FF}0.376  & \cellcolor[HTML]{F5F5FF}14.4\% & \cellcolor[HTML]{F8F2E9}0.618    & \cellcolor[HTML]{F8F2E9}8.7\%  \\
ITER-RETGEN                                  & \cellcolor[HTML]{FCF0EF}0.366       & \cellcolor[HTML]{FCF0EF}0.484 & \cellcolor[HTML]{FCF0EF}8.2\%  & \cellcolor[HTML]{F5F5FF}0.252       & \cellcolor[HTML]{F5F5FF}0.3551 & \cellcolor[HTML]{F5F5FF}13.5\% & \cellcolor[HTML]{F8F2E9}0.668    & \cellcolor[HTML]{F8F2E9}0.6\%  \\
BlendFilter                                  & \cellcolor[HTML]{FCF0EF}0.396       & \cellcolor[HTML]{FCF0EF}0.527 & \cellcolor[HTML]{FCF0EF}-      & \cellcolor[HTML]{F5F5FF}0.286       & \cellcolor[HTML]{F5F5FF}0.378  & \cellcolor[HTML]{F5F5FF}-      & \cellcolor[HTML]{F8F2E9}0.672    & \cellcolor[HTML]{F8F2E9}-      \\ \bottomrule
\end{tabular}%
}
\vspace{-0.05in}
\end{table*}

\begin{table*}[htb!]
    \centering
    \caption{Performance of {\name} with Qwen-7b as the backbone.}
   \vspace{-0.1in}
    \label{tab:qwen}
    \resizebox{.8\textwidth}{!}{%
    \begin{tabular}{@{}lcccccccc@{}}
    \toprule
    \multicolumn{1}{c}{}                         & \multicolumn{3}{c}{\cellcolor[HTML]{FCF0EF}HotPotQA}                                                 & \multicolumn{3}{c}{\cellcolor[HTML]{F5F5FF}2WikiMultihopQA}                                          & \multicolumn{2}{c}{\cellcolor[HTML]{F8F2E9}StrategyQA}            \\ \cmidrule(l){2-9} 
    \multicolumn{1}{c}{\multirow{-2}{*}{Method}} & \cellcolor[HTML]{FCF0EF}Exact Match & \cellcolor[HTML]{FCF0EF}F1    & \cellcolor[HTML]{FCF0EF}IMP    & \cellcolor[HTML]{F5F5FF}Exact Match & \cellcolor[HTML]{F5F5FF}F1    & \cellcolor[HTML]{F5F5FF}IMP    & \cellcolor[HTML]{F8F2E9}Accuracy & \cellcolor[HTML]{F8F2E9}IMP    \\ \midrule
    \multicolumn{9}{c}{Without Retrieval}                                                                                                                                                                                                                                                                                          \\ \midrule
    Direct                                       & \cellcolor[HTML]{FCF0EF}0.144       & \cellcolor[HTML]{FCF0EF}0.238 & \cellcolor[HTML]{FCF0EF}118.1\% & \cellcolor[HTML]{F5F5FF}0.182       & \cellcolor[HTML]{F5F5FF}0.244 & \cellcolor[HTML]{F5F5FF}31.9\% & \cellcolor[HTML]{F8F2E9}0.630    & \cellcolor[HTML]{F8F2E9}4.1\% \\
    CoT                                          & \cellcolor[HTML]{FCF0EF}0.150       & \cellcolor[HTML]{FCF0EF}0.245 & \cellcolor[HTML]{FCF0EF}109.3\% & \cellcolor[HTML]{F5F5FF}0.180       & \cellcolor[HTML]{F5F5FF}0.246 & \cellcolor[HTML]{F5F5FF}33.3\% & \cellcolor[HTML]{F8F2E9}0.658    & \cellcolor[HTML]{F8F2E9}-0.3\%  \\ \midrule
    \multicolumn{9}{c}{With Retrieval}                                                                                                                                                                                                                                                                                             \\ \midrule
    Direct                                       & \cellcolor[HTML]{FCF0EF}0.180       & \cellcolor[HTML]{FCF0EF}0.310 & \cellcolor[HTML]{FCF0EF}74.4\% & \cellcolor[HTML]{F5F5FF}0.084       & \cellcolor[HTML]{F5F5FF}0.200 & \cellcolor[HTML]{F5F5FF}185.7\% & \cellcolor[HTML]{F8F2E9}0.572    & \cellcolor[HTML]{F8F2E9}14.6\% \\
    CoT                                          & \cellcolor[HTML]{FCF0EF}0.206       & \cellcolor[HTML]{FCF0EF}0.305 & \cellcolor[HTML]{FCF0EF}52.4\% & \cellcolor[HTML]{F5F5FF}0.210       & \cellcolor[HTML]{F5F5FF}0.292 & \cellcolor[HTML]{F5F5FF}14.3\% & \cellcolor[HTML]{F8F2E9}0.604    & \cellcolor[HTML]{F8F2E9}8.6\% \\
    ReAct                                        & \cellcolor[HTML]{FCF0EF}0.142       & \cellcolor[HTML]{FCF0EF}0.239 & \cellcolor[HTML]{FCF0EF}121.1\% & \cellcolor[HTML]{F5F5FF}0.158       & \cellcolor[HTML]{F5F5FF}0.241 & \cellcolor[HTML]{F5F5FF}51.9\%  & \cellcolor[HTML]{F8F2E9}0.592    & \cellcolor[HTML]{F8F2E9}10.8\% \\
    SelfAsk                                      & \cellcolor[HTML]{FCF0EF}0.206       & \cellcolor[HTML]{FCF0EF}0.307 & \cellcolor[HTML]{FCF0EF}52.4\% & \cellcolor[HTML]{F5F5FF}0.106       & \cellcolor[HTML]{F5F5FF}0.154 & \cellcolor[HTML]{F5F5FF}126.4\% & \cellcolor[HTML]{F8F2E9}0.596    & \cellcolor[HTML]{F8F2E9}10.1\%   \\
    ITER-RETGEN                                  & \cellcolor[HTML]{FCF0EF}0.244       & \cellcolor[HTML]{FCF0EF}0.364 & \cellcolor[HTML]{FCF0EF}28.7\% & \cellcolor[HTML]{F5F5FF}0.200       & \cellcolor[HTML]{F5F5FF}0.297 & \cellcolor[HTML]{F5F5FF}20.0\% & \cellcolor[HTML]{F8F2E9}0.612    & \cellcolor[HTML]{F8F2E9}7.2\%  \\
    BlendFilter                                  & \cellcolor[HTML]{FCF0EF}0.314       & \cellcolor[HTML]{FCF0EF}0.442 & \cellcolor[HTML]{FCF0EF}-      & \cellcolor[HTML]{F5F5FF}0.240       & \cellcolor[HTML]{F5F5FF}0.312 & \cellcolor[HTML]{F5F5FF}-      & \cellcolor[HTML]{F8F2E9}0.656    & \cellcolor[HTML]{F8F2E9}-      \\ \bottomrule
    \end{tabular}%
    }
    \vspace{-0.05in}
    \end{table*}

The performance results in the tables demonstrate that our proposed {\name} consistently achieves substantial improvements over the baselines across different backbones and datasets. Remarkably, our {\name} model achieves average performance improvements of 9.7\%, 7.4\%, and 14.2\% when using GPT3.5-turbo-Instruct, Vicuna 1.5-13b, and Qwen-7b as backbones, respectively. These results demonstrate the effectiveness of our proposed {\name} in enhancing retrieval-augmented generation performance and its ability to generalize across various backbones.

It is worth noting that mere retrieval does not consistently enhance accuracy. For instance, when comparing CoT with retrieval and CoT without retrieval using GPT3.5-turbo-Instruct on 2WikiMultihopQA (as shown in Table~\ref{tab:gpt-turbo}), CoT without retrieval exhibits a higher Exact Match score than CoT with retrieval. This observation suggests that the retrieved knowledge documents may include unrelated information, which can lead to misleading the LLM. This observation aligns with one of our underlying motivations.

\begin{figure}[htb!] 
\vspace{-0.15in}
\centering    

\subfigure[ColBERT v2] {  
\includegraphics[width=0.46\columnwidth,trim=0 60 0 60,clip]{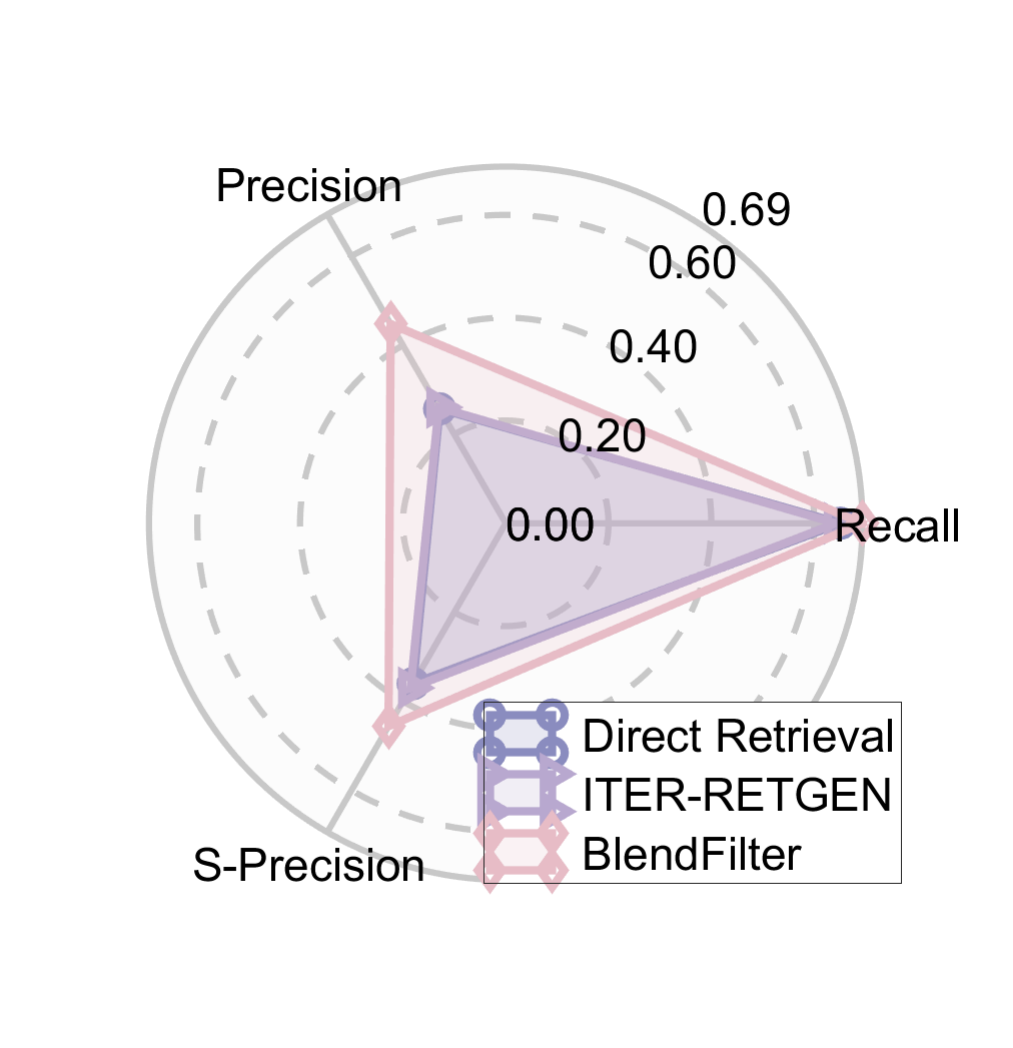}  
\captionsetup{belowskip=-1in}
}     
\subfigure[BM25] {     
\includegraphics[width=0.46\columnwidth,trim=0 60 0 60,clip]{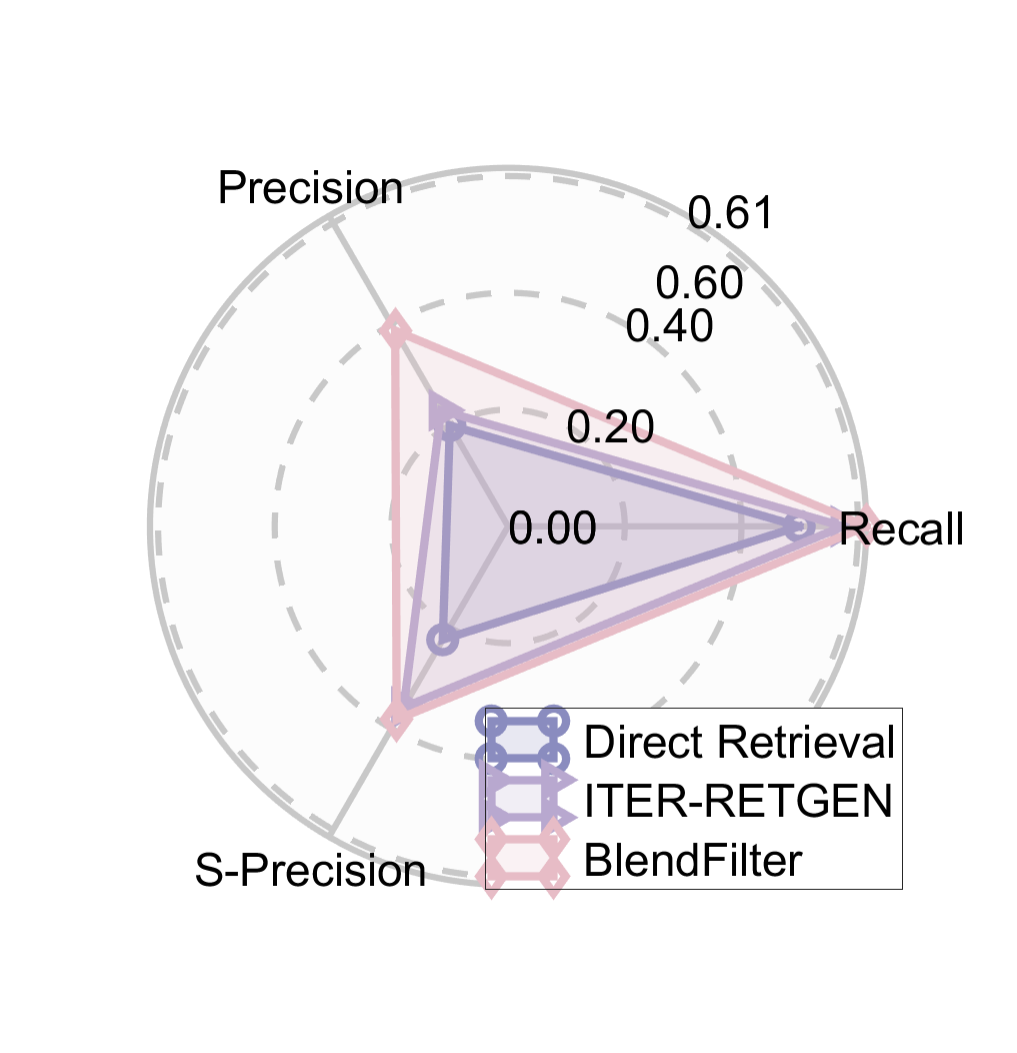}     
\captionsetup{belowskip=-1in}
}
\vspace{-0.15in}
\caption{Retrieval performance after knowledge filtering with GPT3.5-turbo-Instruct on HotPotQA.} 
\label{fig:knowledge_filter}
\vspace{-0.05in}
\end{figure}


\begin{table}[htb!]
\vspace{-0.05in}
\centering
\caption{Performance of {\name} with GPT3.5-turbo-Instruct and BM25 on HotPotQA.}
\vspace{-0.1in}
\label{tab:bm25}
\begin{tabular}{@{}lcc@{}}
\toprule
Method          & Exact Match   & F1      \\ \midrule
\multicolumn{3}{c}{Without Retrieval}     \\ \midrule
Direct          & 0.304         & 0.410   \\
CoT             & 0.302         & 0.432   \\ \midrule
\multicolumn{3}{c}{With Retrieval (BM25)} \\ \midrule
Direct          & 0.342         & 0.462   \\
CoT             & 0.348         & 0.470   \\
ReAct           & 0.280         & 0.371   \\
SelfAsk         & 0.290         & 0.393   \\
ITER-RETGEN     & 0.356         & 0.488   \\
BlendFilter     & 0.420         & 0.547   \\ \bottomrule
\end{tabular}%
\vspace{-0.15in}
\end{table}

\subsection{Combining with BM25}
In this section, we utilize BM25~\cite{10.1016/S0306-4573(00)00016-9}, a widely-used sparse retriever, to explore \textbf{RQ2} on the HotPotQA dataset. The results are shown in Table~\ref{tab:bm25}. When comparing the results in Table~\ref{tab:bm25} with those in Table~\ref{tab:gpt-turbo}, it becomes evident that utilizing ColBERT v2, a dense retriever, yields superior performance compared to BM25. Dense retrievers prove more effective in capturing semantic similarities between questions and documents, especially for complex queries. Moreover, our proposed {\name} consistently outperforms the baselines when BM25 serves as the retriever as well. The proposed {\name} achieves an improvement of approximately 18\%, surpassing the performance when ColBERT v2 is employed as the retriever, in comparison to the baseline models. One potential explanation is that BM25 lacks the potency of ColBERT v2, making the application of query blending to ensure the explicit inclusion of keywords in queries a more crucial factor. This highlights the effectiveness of our proposed {\name} across different retrievers.

\subsection{Effectiveness for Retrieval}
In this section, we address \textbf{RQ3} by computing Precision, Recall, and S-Precision values after conducting knowledge filtering with GPT3.5-turbo-Instruct on the HotPotQA dataset. Results are presented in Figure~\ref{fig:knowledge_filter}. As indicated in Fig.~\ref{fig:knowledge_filter}, the proposed {\name} leads to a substantial improvement in retrieval performance. In both ColBERT v2 and BM25 scenarios, the proposed {\name} demonstrates superior retrieval accuracy compared to direct retrieval and ITER-RETGEN (multi-hop retrieval). Furthermore, when comparing the Recall between ITER-RETGEN and {\name}, it becomes evident that the proposed query blending is effective. This illustrates that combining three queries can recall a greater number of related documents. When comparing the Precision and S-Precision of the baselines with those of {\name}, we observe that the proposed knowledge filtering effectively eliminates unrelated documents.

\subsection{Effectiveness of Different Queries}
In this section, we investigate how performance changes when removing specific queries from the query blending module, addressing \textbf{RQ4}. The results are shown in Table~\ref{tab:query}. According to Table~\ref{tab:query}, it is evident that removing any query from the query blending process results in thedegradation in model performance. This demonstrates the importance of the original query, the externally augmented query, and the internally augmented query in the answer generation process. Additionally, we can find the internal knowledge-augmented query plays a more important role when BM25 is employed. One possible explanation is that when BM25 is used, the retrieval accuracy is not as robust as that of a dense retriever. Consequently, the externally augmented query may still miss some information. This highlights the importance of complementing it with internal knowledge augmentation.
\begin{table}[htb!]
\centering
\caption{Performance of {\name} without different queries with GPT3.5-turbo-Instruct on HotPotQA.}
\vspace{-0.1in}
\label{tab:query}
\begin{tabular}{@{}lcc@{}}
\toprule
Method         & Exact Match & F1    \\ \midrule
\multicolumn{3}{c}{Dense Retriever (ColBERT v2)}  \\ \midrule
BlendFilter    & 0.508       & 0.624 \\
\qquad w/o $q$        & 0.476       & 0.604 \\
\qquad w/o $q_{ex}$   & 0.442       & 0.565 \\
\qquad w/o $q_{in}$   & 0.496       & 0.613 \\ \midrule
\multicolumn{3}{c}{Sparse Retriever (BM25)} \\ \midrule
BlendFilter    & 0.420       & 0.547 \\
\qquad w/o $q$        & 0.410       & 0.532 \\
\qquad w/o $q_{ex}$   & 0.388       & 0.506 \\
\qquad w/o $q_{in}$   & 0.398       & 0.514 \\ \bottomrule
\end{tabular}%
\vspace{-0.1in}
\end{table}

\subsection{Number of Retrieved Documents}
\begin{figure}[htb!] 
\vspace{-0.1in}
\centering        
\includegraphics[width=1\columnwidth]{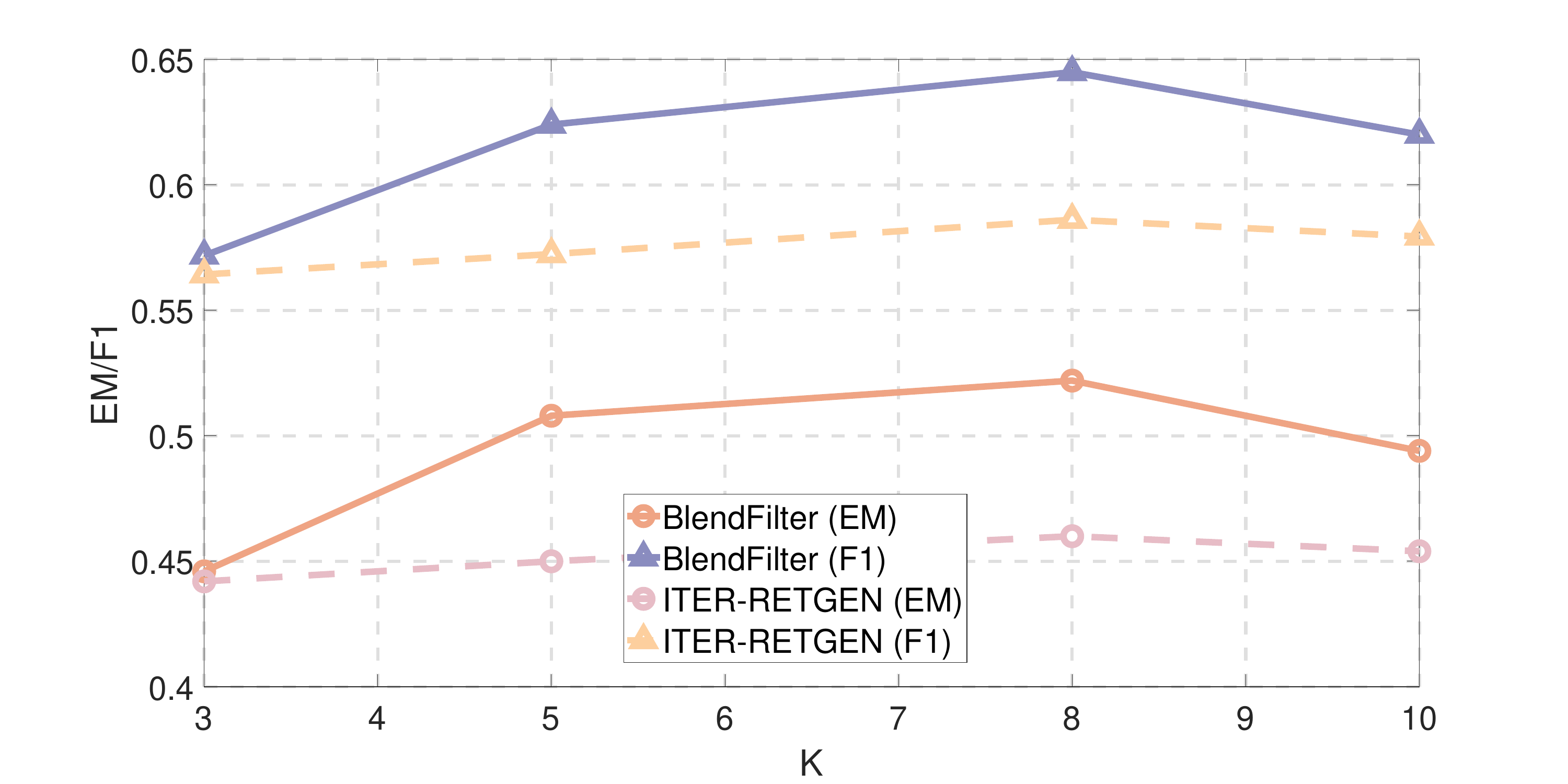} 
\vspace{-0.1in}   
\caption{Performance with respect to different $K$ values on HotPotQA with GPT3.5-turbo-Instruct.} 
\label{fig:num_knowledge}
\end{figure}

\vspace{-0.05in}
In this section, we explore how the model's performance varies when employing different numbers of retrieved documents ($K$), addressing \textbf{RQ5}. The results are presented in Fig.~\ref{fig:num_knowledge}. Based on Fig.~\ref{fig:num_knowledge}, it can be observed that as the value of $K$ is increased, the performance of both ITER-RETGEN and {\name} initially improves and then experiences a slight decline. This indicates that increasing the number of retrieved knowledge documents appropriately can enhance model performance. Notably, it is evident that increasing the value of $K$ from 3 to 8 leads to a substantial improvement in the performance of {\name}, while ITER-RETGEN exhibits only marginal performance gains. One possible explanation is that {\name} incorporates knowledge filtering, effectively eliminating most unrelated knowledge, whereas ITER-RETGEN lacks this filtering mechanism and incorporates a significant amount of noise knowledge.

\subsection{Sampling Times}
\begin{figure}[htb!]
\centering        
\includegraphics[width=1\columnwidth]{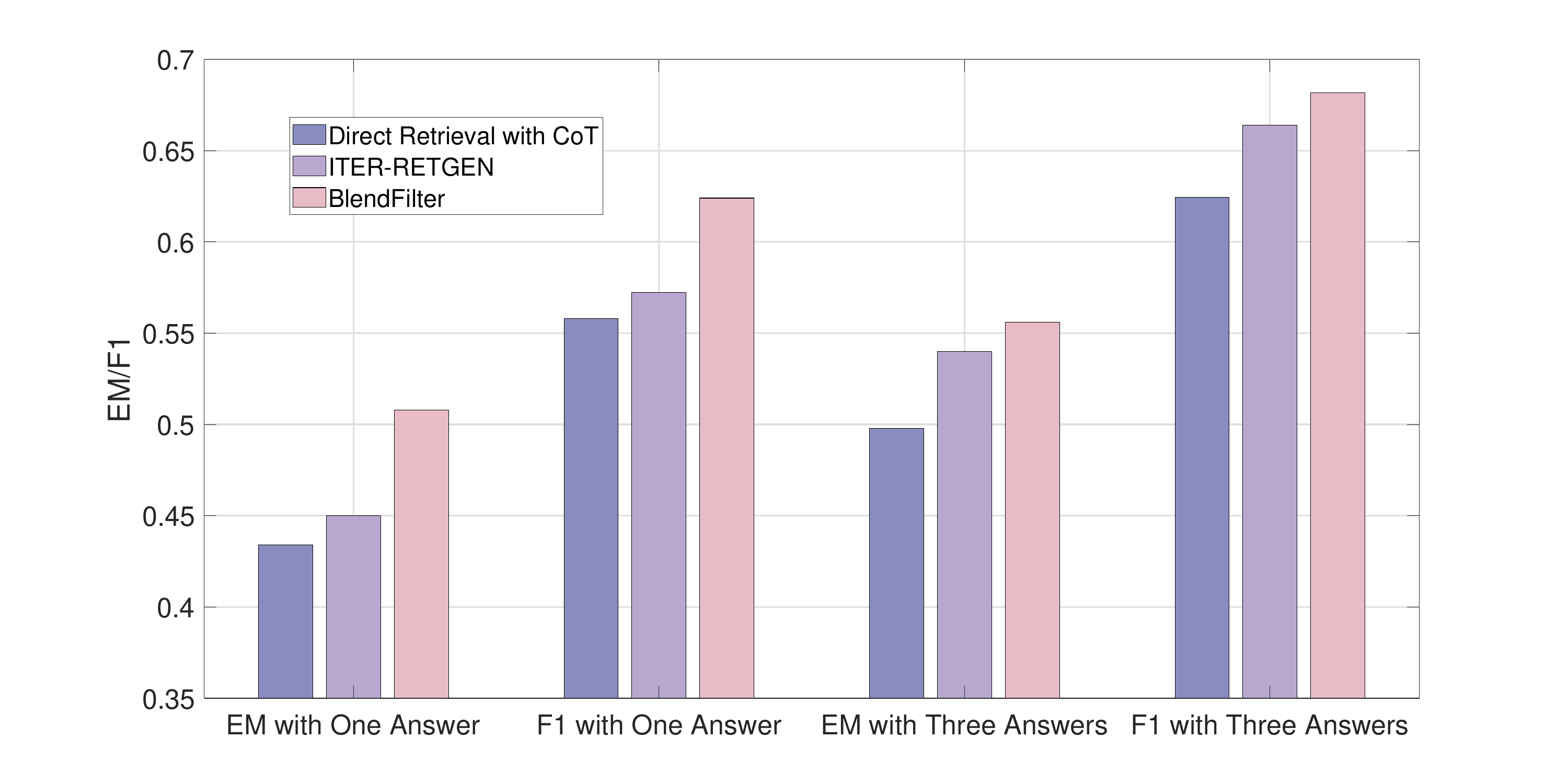}   
\vspace{-0.1in}
\caption{Performance of models with multiple answer sampling on HotPotQA with GPT3.5-turbo-Instruct. For three answers, if one of the answers is correct, its EM will be 1, and the F1 score is the highest one of the three answers.} 
\label{fig:num_answers}
\vspace{-0.25in}
\end{figure}

In this section, we employ various sampling temperatures for the GPT3.5-turbo-Instruct, specifically $\textit{top\_p}=0, 0.5, 1$, and sample one answer under each temperature setting on HotPotQA dataset to address \textbf{RQ6}. The results are shown in Fig.~\ref{fig:num_answers}. Based on Fig.~\ref{fig:num_answers}, it is evident that our proposed {\name} consistently outperforms the baselines, whether sampling a single answer or multiple answers. Furthermore, when three answers are sampled, all methods exhibit improvements, albeit the improvements in the case of {\name} are notably smaller compared to the other baseline methods. This observation demonstrates that when provided with more opportunities to answer, all these models tend to have a higher probability of answering correctly, whereas our proposed {\name} exhibits lower variance.

\subsection{Case Study}
In this section, we show a concrete example in Fig.~\ref{fig:case_study} in the appendix to show how the proposed {\name} works. This example is taken from HotPotQA dataset and we feed it to GPT3.5-turbo-Instruct. The original question is "\textit{superMansion starred the actress who had a recurring role as whom on Workaholics?}". The related knowledge includes the \textit{SuperMasion} document and \textit{Jillian Bell} document. From Fig.~\ref{fig:case_study}, we can find both the original query and external knowledge-augmented query retrieved knowledge consists of one correct document \textit{SuperMasion}. Additionally, the internal knowledge-augmented query retrieves another correct knowledge document \textit{Jillian Bell}. This demonstrates the necessity of combining these three queries to retrieve all relevant knowledge documents. Furthermore, following knowledge filtering, our proposed {\name} effectively eliminates all irrelevant documents and provides the correct answer to the question.

\section{Conclusion}

In this paper, we introduce {\name}, a comprehensive framework developed to enhance retrieval-augmented generation within LLMs. Our methodology distinctively incorporates query generation blending and knowledge filtering techniques, effectively tackling the intricacies of complex inputs and significantly reducing noise in retrieved knowledge. The amalgamation of external and internal knowledge augmentation fosters a resilient and all-encompassing retrieval mechanism. Additionally, our innovative self-reliant knowledge filtering module exploits the inherent capabilities of the LLM to refine and purify the retrieved knowledge by eliminating extraneous content. We conducted extensive experiments on three benchmarks, and the results demonstrate that {\name} outperforms state-of-the-art baselines. Moreover, {\name} can be generalized well for different kinds LLMs, including GPT3.5-turbo-Instruct, Vicuna 1.5-13b and Qwen-7b.

\section*{Limitations}
The proposed {\name} framework introduces a hyper-parameter $K$ to control how many documents we need to retrieve, which might require additional effort to tune. Fortunately, we observe that the model performance is not very sensitive to the hyper-parameter and we set it to a fixed value to achieve a good performance in this paper.

\section*{Acknowledgement}
This work is supported in part by the US National Science Foundation under grant NSF IIS-1747614 and NSF IIS-2141037. Any opinions, findings, and conclusions or recommendations expressed in this material are those of the author(s) and do not
necessarily reflect the views of the National Science Foundation.
\bibliography{custom}
\newpage
\appendix

\begin{table*}[htb!]
    \centering
    \caption{The differences between the proposed {\name} and existing methods.}
    \label{tab:diffs}
    \vspace{-0.1in}
    \resizebox{.8\textwidth}{!}{%
    \begin{tabular}{@{}l|c|c|cc|ccc|c@{}}
    \toprule
    \multirow{2}{*}{}        & \multirow{2}{*}{\begin{tabular}[c]{@{}c@{}}Query \\ Decomposition\end{tabular}} & \multirow{2}{*}{\begin{tabular}[c]{@{}c@{}}Query \\ Rewriting\end{tabular}} & \multicolumn{2}{c|}{Query Augmentation}                                                                                                            & \multicolumn{3}{c|}{Knowledge Selection}                                                                                                                                                                   & \multirow{2}{*}{Need Traing}    \\ \cmidrule(lr){4-8}
                             &                                                                                &                                                                             & \multicolumn{1}{c|}{\begin{tabular}[c]{@{}c@{}}External \\ Knowledge\end{tabular}} & \begin{tabular}[c]{@{}c@{}}Internal \\ Knowledge\end{tabular} & \multicolumn{1}{c|}{\begin{tabular}[c]{@{}c@{}}Predicting Before \\ Retrieval\end{tabular}} & \multicolumn{1}{c|}{\begin{tabular}[c]{@{}c@{}}Model \\ Confidence\end{tabular}} & Filtering                 &                             \\ \midrule
    ReAct~\citet{yao2022react}                    & \Checkmark                                                       & --                                                                            & \multicolumn{1}{c|}{--}                                                              & --                                                              & \multicolumn{1}{c|}{--}                                                                       & \multicolumn{1}{c|}{--}                                                            &--                           & \XSolidBrush \\
    \citet{ma2023query}                         &--                                                                                 & \Checkmark                                                   & \multicolumn{1}{c|}{--}                                                              &--                                                               & \multicolumn{1}{c|}{--}                                                                       & \multicolumn{1}{c|}{--}                                                            &--                           & \Checkmark   \\
    \citet{yu2023improving}                         &--                                                                                 &--                                                                             & \multicolumn{1}{c|}{--}                                                              & \Checkmark                                     & \multicolumn{1}{c|}{--}                                                                       & \multicolumn{1}{c|}{\Checkmark}                                   &--                           & \XSolidBrush \\
    ITER-RETGEN~\cite{shao-etal-2023-enhancing}                         &--                                                                                 &--                                                                             & \multicolumn{1}{c|}{\Checkmark}                                     &--                                                               & \multicolumn{1}{c|}{--}                                                                       & \multicolumn{1}{c|}{--}                                                            &--                           & \XSolidBrush \\
    \citet{asai2023self}                         &--                                                                                 &--                                                                             & \multicolumn{1}{c|}{--}                                                              &--                                                               & \multicolumn{1}{c|}{\Checkmark}                                              & \multicolumn{1}{c|}{--}                                                            &--                           & \Checkmark   \\
    \citet{wang2023self}                         &--                                                                                 &--                                                                             & \multicolumn{1}{c|}{--}                                                              &--                                                               & \multicolumn{1}{c|}{\Checkmark}                                              & \multicolumn{1}{c|}{--}                                                            &--                           & \Checkmark   \\
    {\name} &--                                                                                 &--                                                                             & \multicolumn{1}{c|}{\Checkmark}                                     & \Checkmark                                     & \multicolumn{1}{c|}{--}                                                                       & \multicolumn{1}{c|}{--}                                                            & \Checkmark & \XSolidBrush \\ \bottomrule
    \end{tabular}%
    }
    \vspace{-0.1in}
    \end{table*}

\begin{figure}[h!] 
\vspace{-0.1in}
\centering        
\includegraphics[width=0.95\columnwidth]{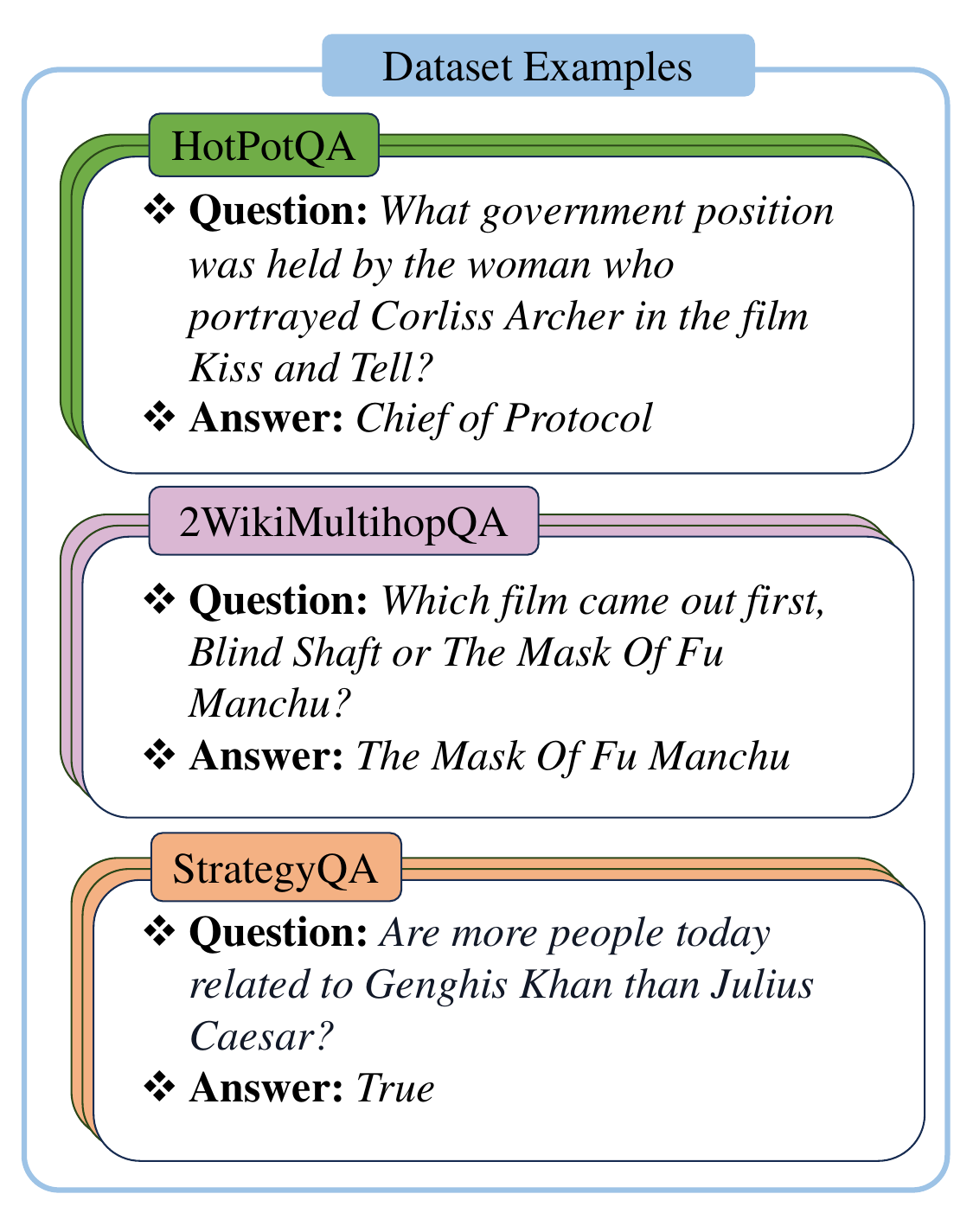} 
\vspace{-0.2in}
\caption{Examples of datasets.} 
\vspace{-0.1in}
\label{fig:data}
\end{figure}
\begin{figure*}[htb!]
\centering        
\includegraphics[width=1.0\textwidth]{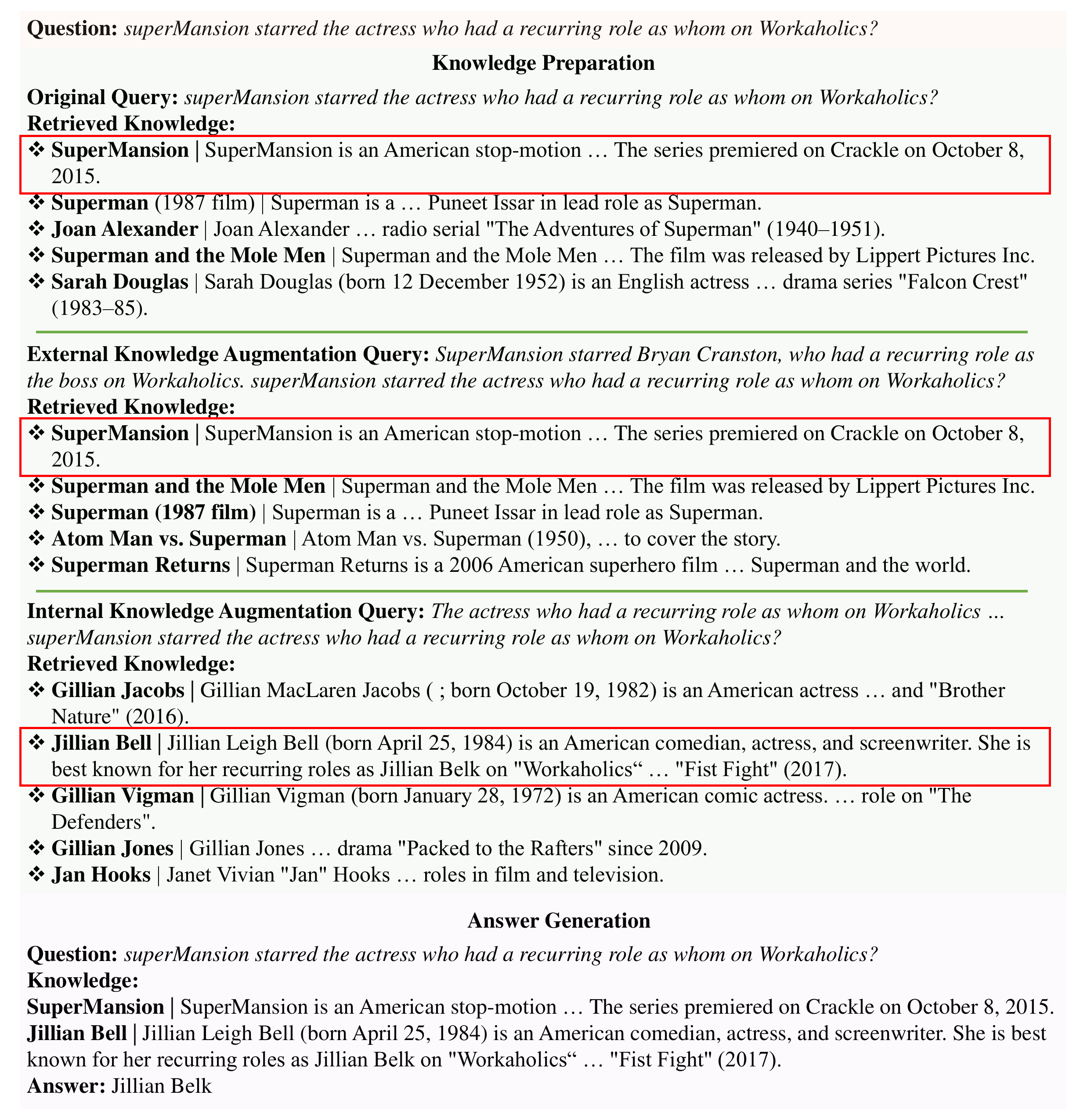}     
\caption{Case study.} 
\label{fig:case_study}
\end{figure*}

\section{Related Work}
We the differences bettwen the proposed {\name} and existing baselines in Table~\ref{tab:diffs}.

\section{Algorithm}
\begin{algorithm}[htb!]
\caption{{\name}}
\label{alg:rag}
\KwInput{An input query $\bm{q}$, a knowledge base $\mathcal{K}$, a retriever $\mathcal{R}(\cdot)$, and a LLM $\mathcal{M}(\cdot)$.}

    

\tcp{query blending}

Direct retrieval by feeding $\bm{q}$ into retriever $\mathcal{R}(\cdot)$;\\
Generate external knowledge-augmented query according to $\bm{a}_{ex}=\mathcal{M}(\bm{a}|\texttt{Prompt}_\texttt{CoT}(\bm{q},\mathcal{K}_{ex}))$ and $\bm{q}_{ex}=\bm{a}_{ex}\|\bm{q}$;\\
Generate internal knowledge-augmented query according to $\bm{a}_{in}=\mathcal{M}(\bm{a}|\texttt{Prompt}(\bm{q}))$ and $\bm{q}_{in}=\bm{a}_{in}\|\bm{q}$;\\

\tcp{Knowledge filtering}
Retrieve knowledge with different queries;\\
Filter retrieved knowledge based on 
\begin{align}
    \notag&\mathcal{K}_{q}=\mathcal{R}(\bm{q},\mathcal{K};K),\\
    \notag&\mathcal{K}_{q_{ex}}=\mathcal{R}(\bm{q}_{ex},\mathcal{K};K),\\
    \notag&\mathcal{K}_{q_{in}}=\mathcal{R}(\bm{q}_{in},\mathcal{K};K);
\end{align}\\
Union filtered knowledge according to $\mathcal{K}_{r}=\mathcal{K}_{q}^{f}\bigcup \mathcal{K}_{q_{ex}}^{f} \bigcup \mathcal{K}_{q_{in}}^{f}$;\\
\tcp{Answer generation}
Generate answer according to $\bm{a}=\mathcal{M}(\bm{a}|\texttt{Prompt}_{\texttt{CoT}}(\bm{q},\mathcal{K}_{r}))$.
\end{algorithm}

\section{Baselines}
We adopt following state-of-the-art baselines to evaluate our proposed {\name}: 
\begin{itemize}[leftmargin=1em]
    \item Direct Prompting~\cite{brown2020language} instructs the LLM to provide direct answers to questions without offering explanations or explicit reasoning steps. We evaluate both Direct Prompting with and without retrieval as our baseline approaches, referring to them as Direct for brevity.
    \item CoT Prompting~\cite{wei2022chain} instructs the LLM to generate answers accompanied by explicit reasoning steps. Similar to Direct Prompting, we evaluate CoT Prompting with and without retrieval, referring to them as CoT in our experiments.
    \item ReAct~\cite{yao2022react} incorporates reasoning, action, and observation steps. The generation process concludes upon reaching the finishing state. The action can involve either generating a query to retrieve knowledge or finalizing the generation. The observation entails the retrieved knowledge documents.
    \item SelfAsk~\cite{press2022measuring} comprises steps for follow-up question generation, retrieval, and answering follow-up questions. Each retrieval operation relies on the generated follow-up questions. When no further follow-up questions are generated, the LLM provides the answer to the original question. We prepend newly retrieved knowledge to the original question following the approach of \citet{yoran-etal-2023-answering}. In the context of this paper, SelfAsk shares similarities with ReAct, albeit differing in the location of retrieved knowledge.
    \item ITER-RETGEN~\cite{shao-etal-2023-enhancing}, a state-of-the-art retrieval-augmented generation method, introduces the iterative augmentation of questions using an external knowledge base and employs knowledge distillation to enhance retriever performance. To ensure a fair comparison, we exclude retrieval training and employ the same retriever as other methods in the case of ITER-RETGEN.
\end{itemize}

\section{Dataset Exmples}

\subsubsection{Implementation Details.}
We evaluate our approach with three different LLMs: GPT3.5-turbo-Instruct$\footnote{\url{https://platform.openai.com/docs/models/gpt-3-5}}$, Vicuna 1.5-13b~\cite{zheng2023judging}, and Qwen-7b~\cite{qwen}. GPT3.5-turbo-Instruct is a refined version of InstructGPT~\cite{ouyang2022training}, Vicuna 1.5-13b is trained based on Llama 2~\cite{touvron2023llama2} continually, and Qwen-7b is a Transformer-based model trained from scratch. Vicuna 1.5-13b and Qwen-7b are open-source models. We utilize the state-of-the-art efficient retrieval method ColBERT v2~\cite{santhanam2022colbertv2} as the retriever implemented by \citet{khattab2022demonstrate,khattab2023dspy} which applies quantization to accelerate approximate nearest neighbor search. We conduct experiments using Vicuna 1.5-13b with vLLM~\citet{kwon2023efficient} and Qwen-7b with Transformers~\cite{wolf-etal-2020-transformers}, respectively. The knowledge base we employ is the collection of Wikipedia abstracts dumped in 2017~\cite{khattab2023dspy}. In all experiments, we utilize a 3-shot in-context learning setting following the approach of \citet{shao-etal-2023-enhancing}. The value of $k$ is set to 5 for all methods. The detailed prompts are provided in the Appendix.

\section{Case Study}
We show an example about how the proposed {\name} works in Fig.~\ref{fig:case_study}.

\newpage

\section{Prompt}
In this section, We show the prompt we use on three benchmarks for GPT3.5-turbo-Instruct, including prompts for external knowledge augmentation, internal knowledge augmentation, knowledge filtering, and answer generation. Among them, the prompt for external knowledge augmentation is the same for all datasets.
\newtcolorbox{mybox_ex_hotpotqa}{
  colback=gray!10, 
  colbacktitle=gray!70, 
  coltitle=black, 
  title=Prompt for External Knowledge Augmentation on HotPotQA, 
  fonttitle=\bfseries, 
  colframe=gray!70, 
  rounded corners, 
  boxrule=0.5mm, 
  drop fuzzy shadow 
}

\begin{mybox_ex_hotpotqa}

Answer questions following the given format.\\

Knowledge:\texttt{\{Example\_Knowledge\}}\\
Question:Are It Might Get Loud and Mr. Big both Canadian documentaries?\\
Let's think step by step.\\
Mr. Big is a 2007 documentary which examines the "Mr. Big" undercover methods used by the Royal Canadian Mounted Police. However, It Might Get Loud is a 2008 American documentary film.\\
So the answer is no.\\

Knowledge:\texttt{\{Example\_Knowledge\}}\\
Question:Were László Benedek and Leslie H. Martinson both film directors?\\
Let's think step by step.\\
László Benedek was a Hungarian-born film director and Leslie H. Martinson was an American film director.\\
So the answer is yes.\\

Knowledge:\texttt{\{Example\_Knowledge\}}\\
Question:Lucium was confimed to be an impure sample of yttrium by an English chemist who became the president of what?\\
Let's think step by step.\\
Lucium was confimed to be an impure sample of yttrium by William Crookes. William Crookes is Sir William Crookes. Sir William Crookes became the president of the Society for Psychical Research.\\
So the answer is Society for Psychical Research.\\

Knowledge:\texttt{\{Knowledge\}}\\
Question:\texttt{\{question\}}\\
Let's think step by step.
\end{mybox_ex_hotpotqa}

\newtcolorbox{mybox_in}{
  colback=gray!10, 
  colbacktitle=gray!70, 
  coltitle=black, 
  title=Prompt for Internal Knowledge Augmentation, 
  fonttitle=\bfseries, 
  colframe=gray!70, 
  rounded corners, 
  boxrule=0.5mm, 
  drop fuzzy shadow 
}

\begin{mybox_in}

Please write a passage to answer the question.\\

Question:\texttt{\{question\}}\\
Passage:
\end{mybox_in}

\newtcolorbox{mybox_knowledge_filter}{
  colback=gray!10, 
  colbacktitle=gray!70, 
  coltitle=black, 
  title=Prompt for Knowledge Filtering on HotPotQA and 2WikiMultihopQA, 
  fonttitle=\bfseries, 
  colframe=gray!70, 
  rounded corners, 
  boxrule=0.5mm, 
  drop fuzzy shadow 
}

\begin{mybox_knowledge_filter}

What general topic is Question \texttt{\{question\}} related to?\\
Answer:The topic is related to\\
---------------------------------------------------------------------------------------------
forget your knowledge about \texttt{\{topic\}}. Please only consider the knowledge below.\\ 
knowledge  0 : \texttt{\{Retrieved\_knowledge0\}}\\
knowledge  1 : \texttt{\{Retrieved\_knowledge1\}}\\
knowledge  2 : \texttt{\{Retrieved\_knowledge2\}}\\
knowledge  3 : \texttt{\{Retrieved\_knowledge3\}}\\
knowledge  4 : \texttt{\{Retrieved\_knowledge4\}}\\
Please check the relevance between \texttt{\{question\}} and knowledges 0-4 one by one, remove the irrelevant ones and show me the relevant ones. There may be multiple relevent ones. Please take a deep breath and do it step by step.\\
---------------------------------------------------------------------------------------------
Please check the relevance between the given question and knowledges 0-4 one by one based on the given context. ONLY output the relevant knowledge ids (0-4). There may be multiple relevent ones.\\

Context:\texttt{\{LLM\_Last\_Generated\_Context\}}\\

Question:\texttt{\{question\}}\\

knowledge  0 : \texttt{\{Retrieved\_knowledge0\}}\\
knowledge  1 : \texttt{\{Retrieved\_knowledge1\}}\\
knowledge  2 : \texttt{\{Retrieved\_knowledge2\}}\\
knowledge  3 : \texttt{\{Retrieved\_knowledge3\}}\\
knowledge  4 : \texttt{\{Retrieved\_knowledge4\}}\\

Answer:

\end{mybox_knowledge_filter}

\newtcolorbox{mybox_generation_hotpotqa}{
  colback=gray!10, 
  colbacktitle=gray!70, 
  coltitle=black, 
  title=Prompt for Answer Generation on HotPotQA, 
  fonttitle=\bfseries, 
  colframe=gray!70, 
  rounded corners, 
  boxrule=0.5mm, 
  drop fuzzy shadow 
}

\begin{mybox_generation_hotpotqa}

Answer questions following the given format.\\

Knowledge:\texttt{\{Example\_Knowledge\}}\\
Question:Are It Might Get Loud and Mr. Big both Canadian documentaries?\\
Let's think step by step.\\
Mr. Big is a 2007 documentary which examines the "Mr. Big" undercover methods used by the Royal Canadian Mounted Police. However, It Might Get Loud is a 2008 American documentary film.\\
So the answer is no.\\

Knowledge:\texttt{\{Example\_Knowledge\}}\\
Question:Were László Benedek and Leslie H. Martinson both film directors?\\
Let's think step by step.\\
László Benedek was a Hungarian-born film director and Leslie H. Martinson was an American film director.\\
So the answer is yes.\\

Knowledge:\texttt{\{Example\_Knowledge\}}\\
Question:Lucium was confimed to be an impure sample of yttrium by an English chemist who became the president of what?\\
Let's think step by step.\\
Lucium was confimed to be an impure sample of yttrium by William Crookes. William Crookes is Sir William Crookes. Sir William Crookes became the president of the Society for Psychical Research.\\
So the answer is Society for Psychical Research.\\

Knowledge:\texttt{\{Filtered\_Knowledge\}}\\
Question:\texttt{\{question\}}\\
Let's think step by step.\\
---------------------------------------------------------------------------------------------
Answer the following question based on the given context with one or few words.\\

Context:\texttt{\{LLM\_Last\_Generated\_Context\}}\\
Question:\texttt{\{question\}}\\
Answer:
\end{mybox_generation_hotpotqa}

\newtcolorbox{mybox_ex_2wiki}{
  colback=gray!10, 
  colbacktitle=gray!70, 
  coltitle=black, 
  title=Prompt for External Knowledge Augmentation on 2WikiMultihopQA, 
  fonttitle=\bfseries, 
  colframe=gray!70, 
  rounded corners, 
  boxrule=0.5mm, 
  drop fuzzy shadow 
}

\begin{mybox_ex_2wiki}

Answer questions following the given format.\\

Knowledge:\texttt{\{Example\_Knowledge\}}\\
Question:Do both films The Falcon (Film) and Valentin The Good have the directors from the same country?\\
Let's think step by step.\\
Valentin The Good is directed by Martin Frič. Martin Frič was a Czech film director. The Falcon (Film) is directed by Vatroslav Mimica. Vatroslav Mimica is a Croatian film director. Czech is different from Croatia.\\
So the answer is no.\\

Knowledge:\texttt{\{Example\_Knowledge\}}\\
Question:What nationality is the director of film Wedding Night In Paradise (1950 Film)?\\
Let's think step by step.\\
Wedding Night In Paradise (1950 film) is directed by Géza von Bolváry. Géza von Bolváry was a Hungarian actor, screenwriter and film director.\\
So the answer is Hungarian.\\

Knowledge:\texttt{\{Example\_Knowledge\}}\\
Question:Who is Rhescuporis I (Odrysian)'s paternal grandfather?\\
Let's think step by step.\\
The father of Rhescuporis I (Odrysian) is Cotys III. The father of Cotys III is Raizdos.\\
So the answer is Raizdos.\\

Knowledge:\texttt{\{Knowledge\}}\\
Question:\texttt{\{question\}}\\
Let's think step by step.
\end{mybox_ex_2wiki}

\newtcolorbox{mybox_generation_2wiki}{
  colback=gray!10, 
  colbacktitle=gray!70, 
  coltitle=black, 
  title=Prompt for Answer Generation on 2WikiMultihopQA, 
  fonttitle=\bfseries, 
  colframe=gray!70, 
  rounded corners, 
  boxrule=0.5mm, 
  drop fuzzy shadow 
}

\begin{mybox_generation_2wiki}

Answer questions following the given format.\\

Knowledge:\texttt{\{Example\_Knowledge\}}\\
Question:Do both films The Falcon (Film) and Valentin The Good have the directors from the same country?\\
Let's think step by step.\\
Valentin The Good is directed by Martin Frič. Martin Frič was a Czech film director. The Falcon (Film) is directed by Vatroslav Mimica. Vatroslav Mimica is a Croatian film director. Czech is different from Croatia.\\
So the answer is no.\\

Knowledge:\texttt{\{Example\_Knowledge\}}\\
Question:What nationality is the director of film Wedding Night In Paradise (1950 Film)?\\
Let's think step by step.\\
Wedding Night In Paradise (1950 film) is directed by Géza von Bolváry. Géza von Bolváry was a Hungarian actor, screenwriter and film director.\\
So the answer is Hungarian.\\

Knowledge:\texttt{\{Example\_Knowledge\}}\\
Question:Who is Rhescuporis I (Odrysian)'s paternal grandfather?\\
Let's think step by step.\\
The father of Rhescuporis I (Odrysian) is Cotys III. The father of Cotys III is Raizdos.\\
So the answer is Raizdos.\\

Knowledge:\texttt{\{Filtered\_Knowledge\}}\\
Question:\texttt{\{question\}}\\
Let's think step by step.\\
---------------------------------------------------------------------------------------------
Answer the following question based on the given context with one or few words.\\

Context:\texttt{\{LLM\_Last\_Generated\_Context\}}\\
Question:\texttt{\{question\}}\\
Answer:
\end{mybox_generation_2wiki}

\newtcolorbox{mybox_ex_stg}{
  colback=gray!10, 
  colbacktitle=gray!70, 
  coltitle=black, 
  title=Prompt for External Knowledge Augmentation on StrategyQA, 
  fonttitle=\bfseries, 
  colframe=gray!70, 
  rounded corners, 
  boxrule=0.5mm, 
  drop fuzzy shadow 
}

\begin{mybox_ex_stg}

Answer questions following the given format.\\

Knowledge:\texttt{\{Example\_Knowledge\}}\\
Question:Do people take laxatives because they enjoy diarrhea?\\
Let's think step by step.\\
Laxatives are substances that loosen stools and increase bowel movements. People take laxatives to treat and/or prevent constipation.\\
So the answer is No.\\

Knowledge:\texttt{\{Example\_Knowledge\}}\\
Question:Could Durian cause someone's stomach to feel unwell?\\
Let's think step by step.\\
Durian has a pungent odor that many people describe as being similar to feet and onions. Unpleasant smells can make people feel nauseous.\\
So the answer is Yes.\\

Knowledge:\texttt{\{Example\_Knowledge\}}\\
Question:Did the swallow play a role in a famous film about King Arthur?\\
Let's think step by step.\\
Monty Python and the Holy Grail was a famous film about King Arthur. In Monty Python and the Holy Grail, swallows are mentioned several times.\\
So the answer is Yes.\\

Knowledge:\texttt{\{Knowledge\}}\\
Question:\texttt{\{question\}}\\
Let's think step by step.
\end{mybox_ex_stg}

\newtcolorbox{mybox_knowledge_filter_stg}{
  colback=gray!10, 
  colbacktitle=gray!70, 
  coltitle=black, 
  title=Prompt for Knowledge Filtering on StrategyQA, 
  fonttitle=\bfseries, 
  colframe=gray!70, 
  rounded corners, 
  boxrule=0.5mm, 
  drop fuzzy shadow 
}

\begin{mybox_knowledge_filter_stg}

Please check the relevance between the given question and knowledges 0-4 one by one carefully, remove all the irrelevant ones and only show me the relevant ones. There may be no relevant one.\\

Question:\texttt{\{question\}}\\

knowledge  0 : \texttt{\{Retrieved\_knowledge0\}}\\
knowledge  1 : \texttt{\{Retrieved\_knowledge1\}}\\
knowledge  2 : \texttt{\{Retrieved\_knowledge2\}}\\
knowledge  3 : \texttt{\{Retrieved\_knowledge3\}}\\
knowledge  4 : \texttt{\{Retrieved\_knowledge4\}}\\

Please take a deep breath and do it step by step.\\
---------------------------------------------------------------------------------------------
Please check the relevance between the given question and knowledges 0-4 one by one based on the given context. ONLY output the relevant knowledge ids (0-4). There may be no relevant one.\\

Context:\texttt{\{LLM\_Last\_Generated\_Context\}}\\

Question:\texttt{\{question\}}\\

knowledge  0 : \texttt{\{Retrieved\_knowledge0\}}\\
knowledge  1 : \texttt{\{Retrieved\_knowledge1\}}\\
knowledge  2 : \texttt{\{Retrieved\_knowledge2\}}\\
knowledge  3 : \texttt{\{Retrieved\_knowledge3\}}\\
knowledge  4 : \texttt{\{Retrieved\_knowledge4\}}\\

Answer:

\end{mybox_knowledge_filter_stg}

\newtcolorbox{mybox_generation_stg}{
  colback=gray!10, 
  colbacktitle=gray!70, 
  coltitle=black, 
  title=Prompt for Answer Generation on StrategyQA, 
  fonttitle=\bfseries, 
  colframe=gray!70, 
  rounded corners, 
  boxrule=0.5mm, 
  drop fuzzy shadow 
}

\begin{mybox_generation_stg}

Answer questions following the given format.\\

Knowledge:\texttt{\{Example\_Knowledge\}}\\
Question:Do people take laxatives because they enjoy diarrhea?\\
Let's think step by step.\\
Laxatives are substances that loosen stools and increase bowel movements. People take laxatives to treat and/or prevent constipation.\\
So the answer is No.\\

Knowledge:\texttt{\{Example\_Knowledge\}}\\
Question:Could Durian cause someone's stomach to feel unwell?\\
Let's think step by step.\\
Durian has a pungent odor that many people describe as being similar to feet and onions. Unpleasant smells can make people feel nauseous.\\
So the answer is Yes.\\

Knowledge:\texttt{\{Example\_Knowledge\}}\\
Question:Did the swallow play a role in a famous film about King Arthur?\\
Let's think step by step.\\
Monty Python and the Holy Grail was a famous film about King Arthur. In Monty Python and the Holy Grail, swallows are mentioned several times.\\
So the answer is Yes.\\

Knowledge:\texttt{\{Filtered\_Knowledge\}}\\
Question:\texttt{\{question\}}\\
Let's think step by step.\\
---------------------------------------------------------------------------------------------
Answer the following question based on the given context. The final answer to a question should always be either Yes or No, and NOTHING ELSE.\\

Context:\texttt{\{LLM\_Last\_Generated\_Context\}}\\
Question:\texttt{\{question\}}\\
Answer:
\end{mybox_generation_stg}

\end{document}